\pdfoutput=1

\documentclass{cup-hpl}
\usepackage{multirow} 
\usepackage{makecell}
\usepackage{lipsum}

\usepackage{bm}
\usepackage{threeparttable}
\usepackage{booktabs}

\usepackage{xcolor}
\usepackage[hidelinks]{hyperref}
\newcommand{\tabref}[1]{Table~\textcolor{red}{\hyperref[#1]{\ref*{#1}}}}
\newcommand{\figref}[1]{\textcolor{blue}{\hyperref[#1]{Figure~\ref*{#1}}}}
\newcommand{\refcite}[1]{\textcolor{blue}{\cite{#1}}}
\usepackage{stfloats}
\usepackage{graphicx}
\usepackage{booktabs}
\begin{document}
%\linenumbers
\newtheorem{theorem}{Theorem}

\shorttitle{CFW-GMN}   
\shortauthor{Y.Han et al.}

\title{Confidence-feedback-weighted graph matching network: online-offline laser-induced damage site matching under complex interference}

\author[1]{Yueyue Han}
\author[1]{Guanhua Chen}
\author[1]{Hangcheng Dong}
\author[1]{Kang Zhang}
\author[1]{Fengdong Chen}
\author[2]{Zhitao Peng}
\author[2]{Fa Zeng}
\author[2]{Qihua Zhu}
\author[1]{Guodong Liu \corresp{F. Chen and G. Liu, No. 92 Xidazhi Street, Harbin 150001, China.
                       \email{chenfd@hit.edu.cn (F. Chen), lgd@hit.edu.cn (G. Liu)}}}

\address[1]{School of Instrumentation Science and Engineering, Harbin Institute of Technology, Harbin 150001, China}
\address[2]{Research Center of Laser Fusion, China Academy of Engineering Physics, Mianyang 621900, China}

\begin{abstract} Online inspection images of final optics in high-power laser facilities contain pseudo-damage sites that closely resemble true damage sites. Determining the authenticity of online-detected sites is therefore difficult and requires accurate matching to offline ground-truth sites. However, this matching remains highly challenging due to limited match-discriminative features, local geometric distortions, and numerous distractor sites. Existing matching models mainly suppress distractors implicitly through loss-function supervision. We propose a confidence-feedback-weighted graph matching network that requires only damage-site centroid coordinates as input. It estimates node matchability confidence from each round of matching scores and feeds it back as a reliability weight to guide subsequent edge-feature aggregation, thereby suppressing distractor propagation and enhancing cross-graph discriminability. Within this framework, a geometric consistency constraint calibrates spurious high-confidence matchability estimates, while a hard-example mining loss improves discrimination between structurally similar sites. Experiments on our Complex-Scene dataset show that the proposed method achieves a matching F1-score of 96.36$\%$ with robust and efficient performance.
\end{abstract}


\maketitle
\section{Introduction}
% \cite{bib1} \cite{bib11,bib12,bib13} \refcite{bib1}
On 7 April 2025, the National Ignition Facility (NIF) achieved a record fusion yield of 8.6 MJ in an inertial confinement fusion experiment\refcite{bib1}. Frequent high-energy ignition shots can increase the risk of laser-induced damage (LID) on the exit surface of the final optics assembly (FOA)\refcite{bib2,bib3,bib4}. Damage growth may lead to severe consequences, including transmitted-beam modulation and cascading damage to downstream optics\refcite{bib5}. The Final Optics Damage Inspection (FODI) system\refcite{bib6,bib7,bib8} performs full-aperture imaging of the FOA to monitor the online damage state of the optics, enabling timely and effective control measures and supporting the long-term safe operation of the high-power laser facility.

\begin{figure}[htbp]
\centering
\includegraphics[width=1\linewidth]{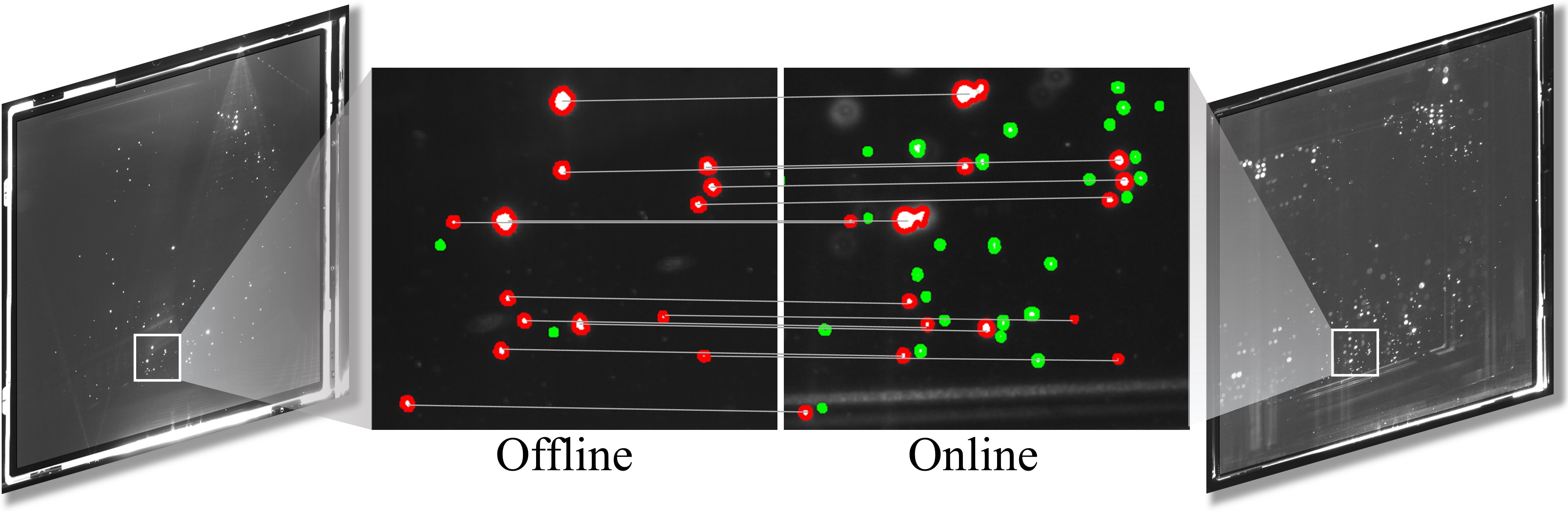}
\caption{Schematic illustration of online-offline damage-site matching. Red regions denote true damage sites with corresponding matches, whereas green regions denote detected damage candidates without true counterparts, referred to as non-corresponding sites.}
\label{fig:off-on}
\end{figure}

Pseudo-damage sites in online inspection images can interfere with online damage-state assessment of the FOA. For the same optic, offline dark-field images of true damage sites can be acquired using an offline inspection platform after dust removal and stray-light suppression\refcite{bib9}. As shown in \figref{fig:off-on}, each finite-area damage region is represented by its centroid point, and correspondences between online-detected and offline damage sites are established based on these centroid locations to verify the authenticity of online damage sites. This matching can also provide large-scale, high-quality labels for training online damage recognition models\refcite{bib10}.

\begin{figure*}[htbp]
\centering
\includegraphics[width=1\textwidth]{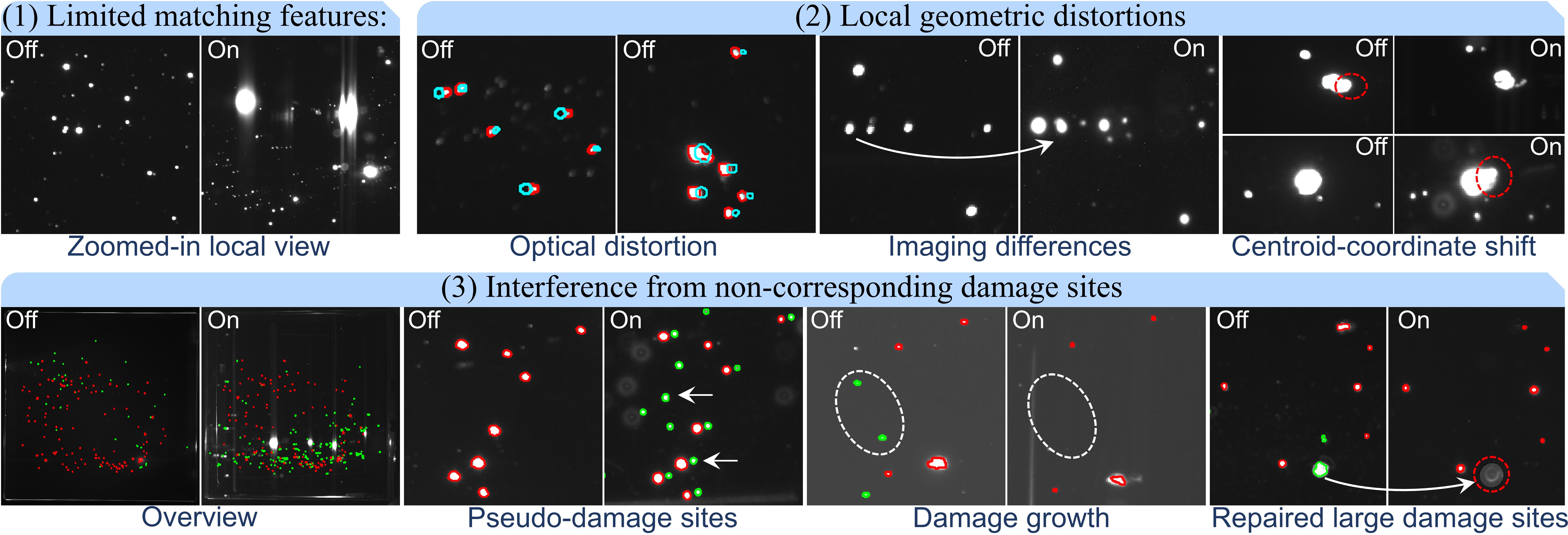}
\caption{Typical challenges in online-offline damage-site matching. Red regions indicate matched true damage sites. Green regions indicate non-corresponding damage sites. Cyan regions denote online damage sites reprojected onto the offline image.}
\label{fig:challenges}
\end{figure*}

This task is considerably more challenging than conventional image matching due to several complex factors, as illustrated in \figref{fig:challenges}. (1) Limited match-discriminative features: Under dark-field side illumination, damage sites provide limited texture and local structural information, while large-area damage sites exhibit central overexposure, resulting in insufficient reliable features for matching. (2) Local geometric distortions: Off-axis tilted imaging of online optics\refcite{bib11,bib12}, optical distortion, platform-dependent imaging differences, and damage growth or repair during asynchronous acquisition\refcite{bib13,bib14} can cause non-uniform shifts in damage-site centroid coordinates and local relative positions, making it difficult to accurately describe the cross-image mapping using a single global homography model. (3) Non-corresponding distractor sites: Stray-light-induced pseudo-damage sites\refcite{bib15} and non-strictly synchronized online-offline acquisition can introduce numerous single-sided non-corresponding damage sites. (4) Limited online-offline paired samples: The Optics Recycle Loop Strategy\refcite{bib16} strictly controls damage evolution, limiting the accumulation of online-offline paired image samples.

Classical local-feature-based matching methods, such as SIFT, ORB, and BRISK\refcite{bib17,bib18,bib19,bib20,bib21,bib22,bib23}, require no training and are generally computationally efficient. However, factor (1) can make it difficult for these methods to extract stable and repeatable local features from damage images, thereby degrading their matching performance. Star identification algorithms\refcite{bib24,bib25}, which are relevant to our task in terms of sparse point-pattern matching, typically exploit local geometric invariants\refcite{bib26,bib27,bib28,bib29,bib30}, such as inter-star distances and included angles, to construct matching constraints or discriminative descriptors for accurate and efficient matching. Nevertheless, factors (2) and (3) may distort or degrade these geometric features. Additional geometric constraints and handcrafted descriptors can partially alleviate this issue, but may also increase computational complexity and storage overhead in large-scale damage-site matching.

\begin{figure}[htbp]
\centering
\includegraphics[width=0.70\linewidth]{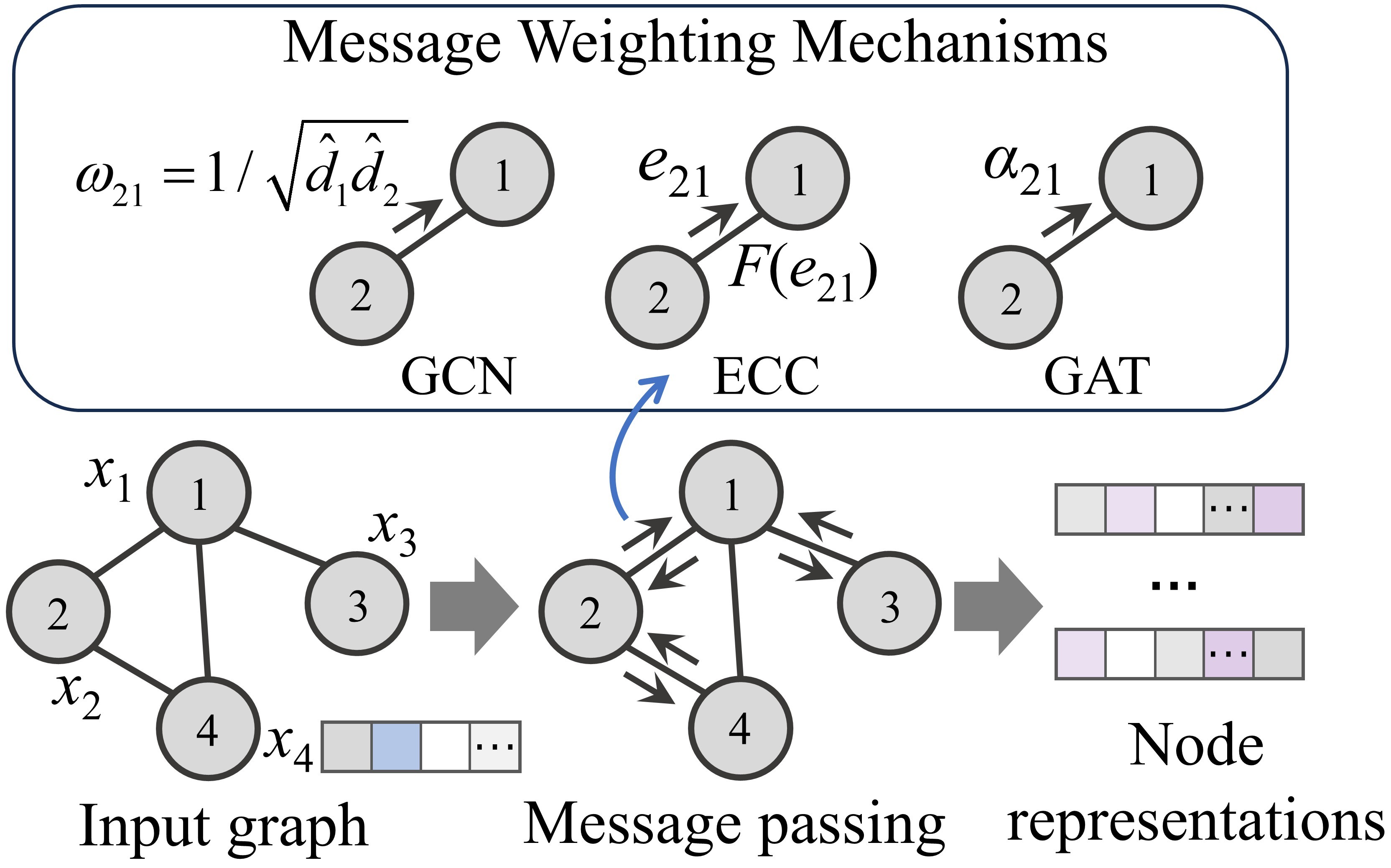}
\caption{Classical message-passing mechanisms of GNNs. $\omega_{21}$, $F(e_{21})$, and $\alpha_{21}$ denote the structural normalization weight in GCN, the edge-conditioned weight in ECC, and the attention weight in GAT, respectively; $e_{21}$ is the edge feature from node 2 to node 1.}
\label{fig:gcn}
\end{figure}

Neural networks reduce reliance on handcrafted constraints by learning data-driven representations. Damage sites exhibit sparse distributions and cross-region spatial dependencies, making them suitable for graph-structured modeling. GNNs\refcite{bib31,bib32,bib33} can therefore learn discriminative features and enable end-to-end matching inference. Graph Convolutional Networks (GCNs)\refcite{bib34}, as representative GNN models, aggregate neighborhood information through graph convolution. However, their predefined aggregation schemes may limit fine-grained modeling of complex node relationships. To address this limitation, Edge-Conditioned Convolution (ECC)\refcite{bib35} incorporates edge attributes to dynamically modulate convolutional weights, while Graph Attention Networks (GATs)\refcite{bib36} use attention mechanisms to adaptively weight neighboring nodes. These methods improve the ability of GNNs to capture complex structural relationships within graphs. In image matching, SuperGlue\refcite{bib37}, LoFTR\refcite{bib38}, and related methods further strengthen the modeling of global cross-image correspondences by introducing self- and cross-attention mechanisms\refcite{bib39,bib40}, leading to improved matching performance. \figref{fig:gcn} illustrates the basic message-passing mechanisms of GCN, ECC, and GAT.

\begin{figure*}[!b]
\centering
\includegraphics[width=1\textwidth]{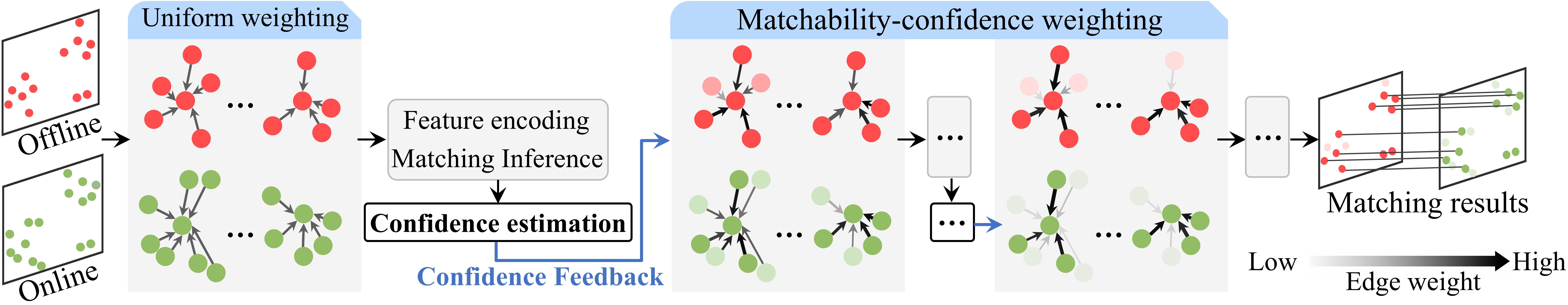}
\caption{Schematic illustration of suppressing interference from non-corresponding sites via the iterative confidence-feedback weighting mechanism in CFW-GMN.}
\label{fig:overview}
\end{figure*}

Existing GNN- and attention-based matching methods often lack an explicit mechanism for suppressing interference from non-corresponding sites. Features from these sites may be propagated during aggregation, weakening true correspondences and reducing matching accuracy. To address this issue, we propose a confidence-feedback-weighted graph matching network (CFW-GMN), which iteratively estimates the matchability confidence of each damage site and feeds it back to the edge-feature aggregation layer as an adaptive weight. As shown in \figref{fig:overview}, this mechanism strengthens information interaction among reliable nodes while suppressing propagation from non-corresponding sites, enabling more discriminative graph representations for accurate and efficient cross-graph damage-site matching.

\section{Methodology}
\begin{figure*}[htbp]
\centering
\includegraphics[width=1\textwidth]{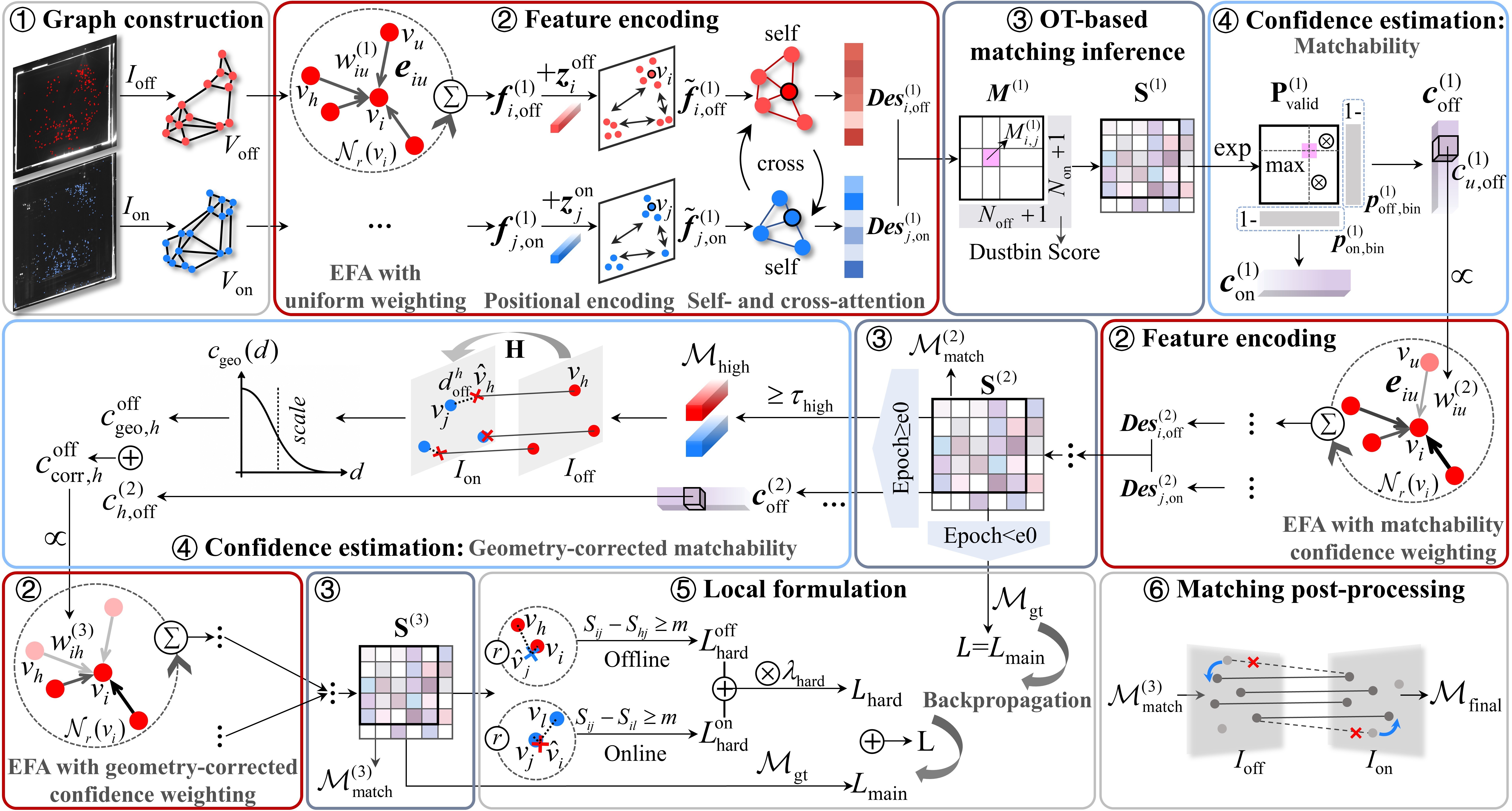}
\caption{The pipeline of CFW-GMN.The variables in the figure are consistent with those defined in the main text.}
\label{fig:pipeline}
\end{figure*}

CFW-GMN is built upon a three-round iterative framework. The first two rounds consist of feature encoding, matching inference, and confidence feedback, while the final round performs feature encoding and matching inference to produce the final correspondences. As shown in \figref{fig:pipeline}, the overall framework comprises six components: graph construction, feature encoding, optimal transport-based matching inference, confidence estimation, loss formulation, and matching post-processing.

\subsection{Graph construction}

The proposed method avoids relying on potentially unstable local appearance features, such as grayscale, texture, and morphological information. Instead, it constructs edge features from relative geometric information between damage sites. Given an offline–online image pair $I_{\mathrm{off}}$ and $I_{\mathrm{on}}$, damage sites are detected using our previously developed CG-Fusion CAM method\refcite{bib41}. Their centroid coordinates are then extracted as the network input and organized into two node sets, $V_{\mathrm{off}}$ and $V_{\mathrm{on}}$, respectively. For each node $v_i$, its local neighborhood is defined as $\mathcal{N}_r(v_i)=\left\{v_u \mid \| \boldsymbol{p}_i - \boldsymbol{p}_u \| \leq r,\ u \neq i\right\}$, where $\boldsymbol{p}_i$ and $\boldsymbol{p}_u$ are the normalized spatial coordinates of $v_i$ and $v_u$. If fewer than $k$ neighboring nodes are found within radius $r$, the neighborhood is supplemented using a $k$-nearest-neighbor strategy. The edge set is then defined as $E=\left\{(v_i,v_u) \mid v_u \in \mathcal{N}_r(v_i)\right\}$, yielding the graph $G=(V,E)$. For each edge $(v_i,v_u)\in E$, we define $\Delta x_{iu}=x_i-x_u$, $\Delta y_{iu}=y_i-y_u$, and $\theta_{iu}=\operatorname{atan2}(\Delta y_{iu}, \Delta x_{iu})$. The corresponding edge feature is represented as $\boldsymbol{e}_{iu}=[\Delta x_{iu},\Delta y_{iu},\sin \theta_{iu},\cos \theta_{iu}]$.

\subsection{Feature encoding}
In CFW-GMN, feature encoding is performed iteratively over three stages. Each stage consists of edge-feature aggregation (EFA), positional encoding, and local self- and cross-attention. The stages share the same overall structure but differ in the EFA weighting strategy: the first stage uses uniform weighting, whereas the second and third stages use matchability-confidence weighting.

\subsubsection{Weighted edge feature aggregation}\mbox{}\\
Given the constructed graph, CFW-GMN updates node representations by aggregating neighborhood edge features. Taking the offline graph as an example, a learnable mapping function $\phi(\cdot)$ first projects the edge feature $\boldsymbol{e}_{iu}$ into the same feature space as the node representation. At the $n$-th EFA stage, the normalized weight $w_{iu}^{(n)}$ modulates the contribution of each neighboring edge, and the weighted aggregation produces a local geometric feature for node $v_i$:

\begin{equation}
\boldsymbol{f}_{i, \mathrm{off}}^{(n)}=\sum_{u \in N_{r}\left(v_{i}\right)} w_{i u}^{(n)} \phi\left(\boldsymbol{e}_{i u}\right).
\end{equation}

To enhance the representation of local geometric context, EFA is implemented with two stacked message-passing layers. The first layer aggregates edge features from the 1-hop neighborhood, whereas the second layer further incorporates structural information from the 2-hop neighborhood, forming a more comprehensive representation of local geometric relationships. All EFA stages share the same parameters to ensure consistent feature learning.

\paragraph{Uniform weighting.} In the first EFA stage, matchability confidence is not yet available, and the model cannot distinguish reliable nodes. Therefore, all neighboring edge features are assigned uniform weights for local geometric aggregation. This stage encodes neighborhood geometric relationships, described by relative displacement and orientation information, into node representations and provides initial features for subsequent matching inference. The uniform aggregation weight is defined as:
\begin{equation}
w_{iu}^{(1)}=\frac{1}{\left|\mathcal{N}_{r}\left(v_{i}\right)\right|}.
\end{equation}

\paragraph{Matchability-confidence weighting.} After the initial matching inference, a matchability confidence score is estimated for each damage site to quantify the likelihood of a true counterpart, and is subsequently used to adaptively modulate neighboring edge features during aggregation. Through this feedback mechanism from cross-graph matching information to intra-graph feature aggregation, the propagation of local geometric information can be explicitly regulated by node reliability. For the offline graph, the confidence values for the second and third EFA stages, $c_{u,\mathrm{off}}^{(1)}$ and $c_{\mathrm{corr},u}^{\mathrm{off}}$, are obtained from different stages of matching inference, with $c_{\mathrm{corr},u}^{\mathrm{off}}$ further corrected based on geometric consistency. The detailed estimation procedure is provided in Section 2.4. For node $v_i$, the aggregation weights associated with its neighboring node $v_u$ in the second and third aggregation rounds are respectively given by:

\begin{equation}
w_{i u}^{(2)}=\underset{v_{u} \in \mathcal{N}_{r}\left(v_{i}\right)}{\operatorname{softmax}}\left(\eta \log \left(c_{u, \mathrm{off }}^{(1)}+\varepsilon\right)\right),
\end{equation}

\begin{equation}
w_{i u}^{(3)}=\underset{v_{\mu} \in \mathcal{N}_{r}\left(v_{i}\right)}{\operatorname{softmax}}\left(\eta \log \left(c_{\text {corr}, u}^{\text{off}}+\varepsilon\right)\right).
\end{equation}

 Here, $\eta$ is the modulation coefficient, and $\varepsilon$ is a small positive constant used to ensure numerical stability.

\begin{figure}[ht]
\centering
\includegraphics[width=1\linewidth]{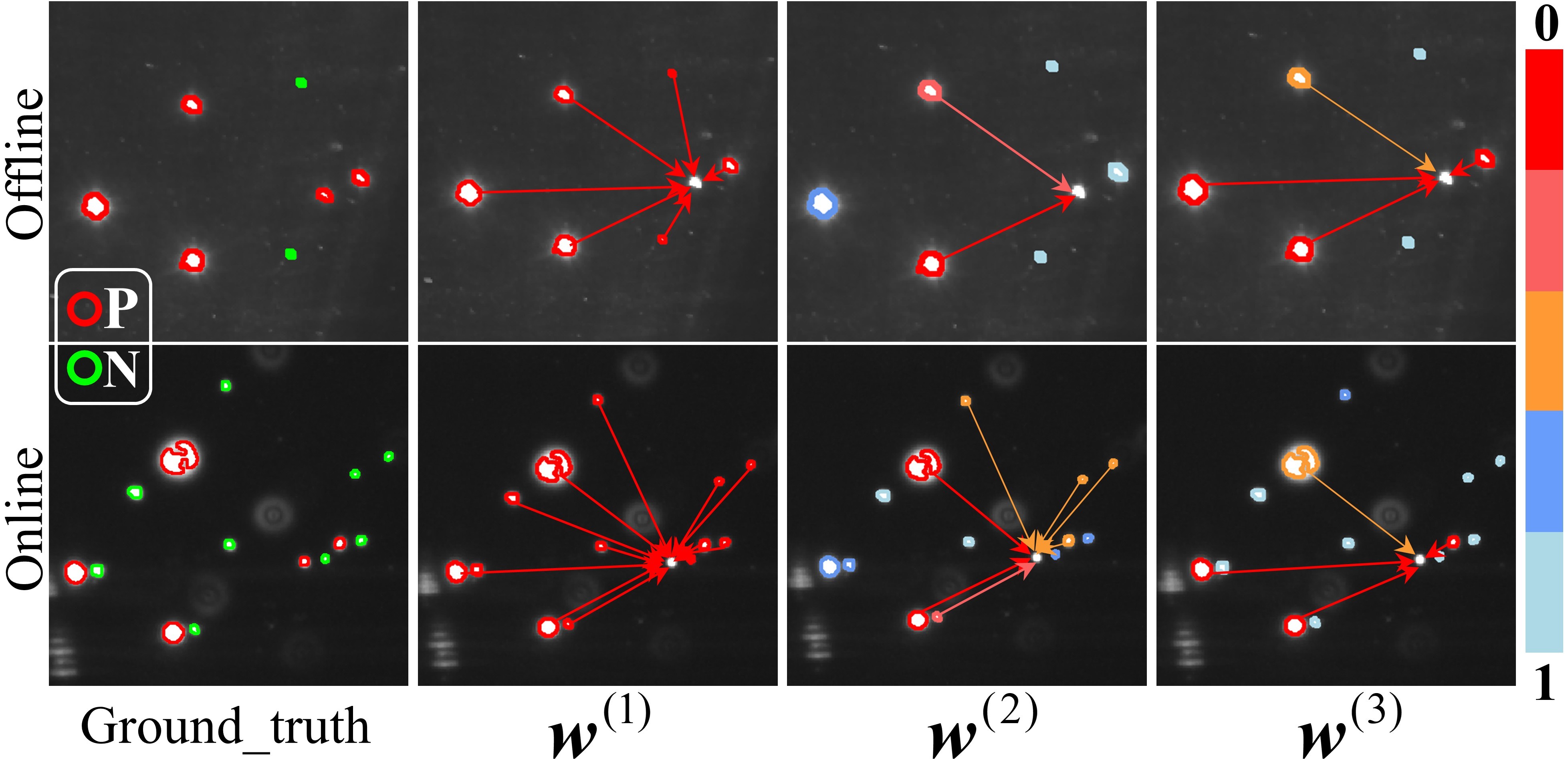}
\caption{Visualization of edge-feature aggregation weights. Red regions indicate matched true damage sites. Green regions indicate non-corresponding damage sites.}
\label{fig:weights}
\end{figure}

\figref{fig:weights} illustrates the evolution of edge-feature aggregation weights over the three EFA iterations. In the initial stage, local geometric information is aggregated from neighboring nodes with uniform weights. After the introduction of matchability confidence, the weights gradually concentrate on reliable nodes that are more likely to have true counterparts. With geometric-consistency correction, the contributions of nodes deviating from the global projective relationship are suppressed, whereas those of geometrically consistent nodes are enhanced. The EFA module in CFW-GMN exhibits a mechanism analogous to manual matching: when matching a damage site, the model uses the local geometric relationship between the site and its neighbors to identify structurally similar candidates in the other image; through repeated cross-image interaction and comparison, it progressively focuses its attention on nodes that better satisfy this geometric relationship while effectively ignoring distractors unlikely to form reliable correspondences, thereby improving matching reliability. 

\subsubsection{Positional encoding}\mbox{}\\
After EFA, the node representation $\mathbf{f}_i^{(n)}$ primarily encodes local geometric relationships characterized by relative distances and angles. However, nodes at different spatial locations may share similar local structures, leading to feature ambiguity. To enhance the global positional awareness of node representations, this study introduces positional encoding based on random Fourier features\refcite{bib42,bib43}. Specifically, the random Fourier feature mapping $\phi_{\mathrm{RFF}}(\cdot)$ is adopted to encode the node coordinate $\mathbf{p}_i$ as:

\begin{equation}
\boldsymbol{z}_{i}=\phi_{\mathrm{REF}}\left(\boldsymbol{p}_{i}\right).
\end{equation}

The RFF mapping $\phi_{\mathrm{RFF}}(\cdot)$ provides a continuous positional encoding with local smoothness, allowing spatially neighboring nodes to obtain similar encodings. It introduces global positional information while preserving the local geometric similarity captured by $\mathbf{f}_i^{(n)}$. The node representation is fused with the positional encoding as $\tilde{\mathbf{f}}_i^{(n)} = \mathbf{f}_i^{(n)} + \mathbf{z}_i$, enabling the fused representation to capture both local geometric structures and global positional information and thereby alleviating mismatches caused by similar local structures.

\subsubsection{ Local self- and cross-attention}
\begin{figure}[ht]
\centering
\includegraphics[width=1\linewidth]{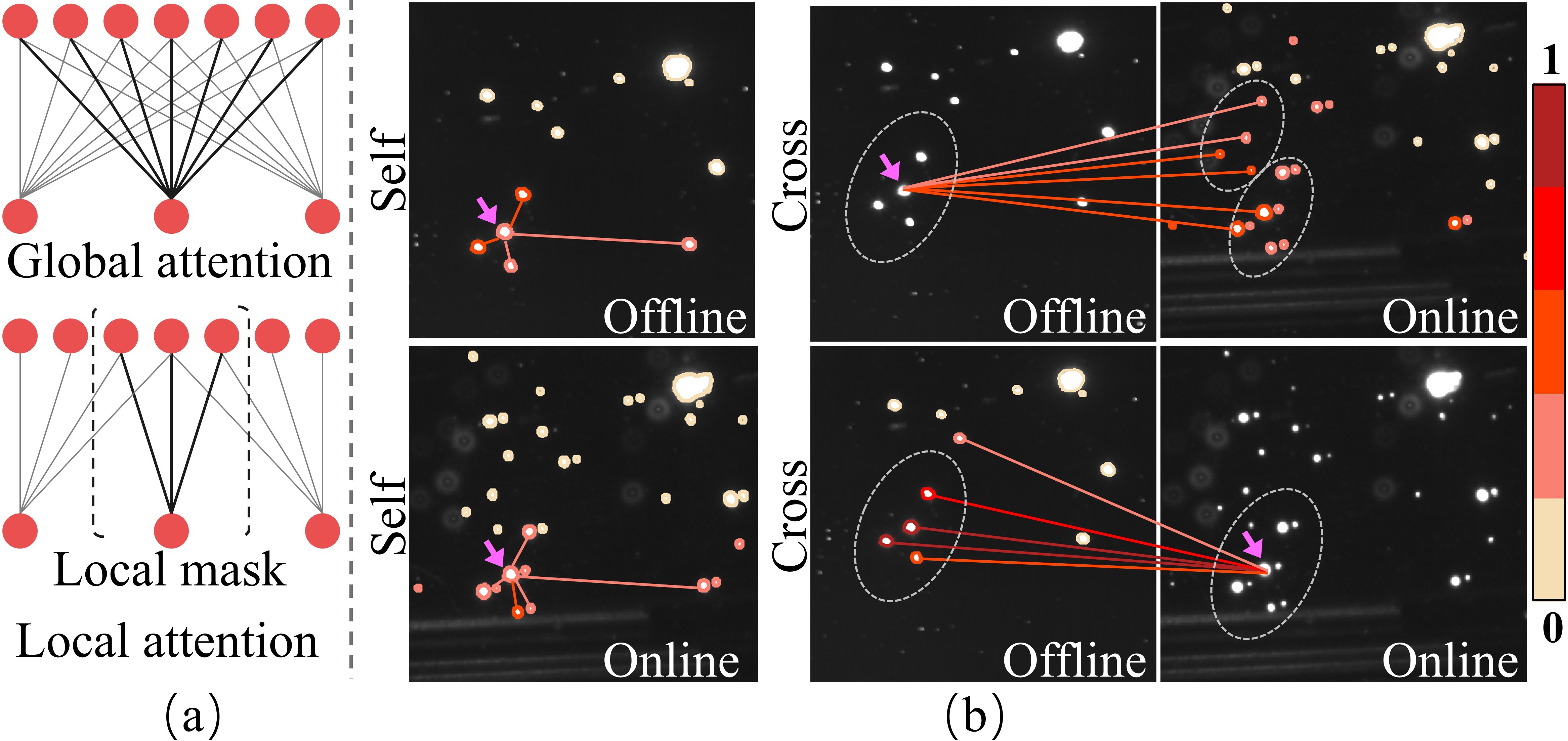}
\caption{(a) Illustration of the receptive fields of global and local attention mechanisms. (b) Visualization of the attention weights generated by local self-attention and cross-attention.}
\label{fig:attention}
\end{figure}

\mbox{}\\
To further enhance the context-aware modeling of node representations and capture discriminative inter-graph interactions, this study introduces multi-head self- and cross-attention mechanisms. In addition, a local attention mask is designed, as shown in \figref{fig:attention}, to restrict the attention scope, suppress irrelevant long-range interactions, and emphasize local structural dependencies. Specifically, the local attention mask $\mathbf{M}_{\mathrm{local}}$ is constructed based on the spatial distance between nodes. When the inter-node distance satisfies $d_{ij} \leq d_{\mathrm{th}}$, the corresponding attention entry is kept valid; otherwise, it is masked during attention computation. The local attention output of the $m$-th attention head is defined as:

\begin{equation}
\mathbf{D e s}_{m}=\left[\operatorname{softmax}\left(\frac{\mathbf{Q}_{m} \mathbf{K}_{m}^{\mathrm{T}}}{\sqrt{d_{k}}}\right) \odot \mathbf{M a s k}_{\text {local }}\right] \mathbf{V}_{m},
\end{equation}

where $\mathbf{Q}_m$, $\mathbf{K}_m$, and $\mathbf{V}_m$ denote the query, key, and value matrices of the $m$-th attention head, respectively. $\sqrt{d_k}$ is the scaling factor, and $d_k$ represents the feature dimension of each attention head. In this module, the self- and cross-attention layers are alternately stacked twice. Subsequently, the outputs of the $h$ attention heads are concatenated and projected into the final $d$-dimensional feature space through the linear transformation matrix $\mathbf{W}_o$, yielding the node descriptor after the $n$-th encoding layer:

\begin{equation}
\mathbf{Des}^{(n)}
=
\operatorname{Concat}
\left(
\mathbf{Des}_1,
\mathbf{Des}_2,
\ldots,
\mathbf{Des}_h
\right)
\mathbf{W}_o ,
\end{equation}

\subsection{Optimal transport-based matching inference}
In CFW-GMN, this module is also executed three times. Note that only the second and third iterations perform matching inference and output predicted correspondences, whereas the first iteration only applies optimal transport to generate matching scores for estimating node matchability confidence.

\paragraph{Optimal transport(OT).}
After obtaining the node descriptors, we infer inter-graph correspondences using descriptor similarities. To address one-to-many and many-to-one mismatches caused by independent matching and to handle nodes without valid correspondences, we introduce a log-domain Sinkhorn optimal transport\refcite{bib44,bib45} matching method with a dustbin mechanism. It generates a globally consistent soft matching matrix under assignment constraints.

Specifically, the similarity matrix $\mathbf{M}^{(n)}$ is constructed from the node descriptors $\mathbf{Des}_{\mathrm{off}}^{(n)}$ and $\mathbf{Des}_{\mathrm{on}}^{(n)}$ of the offline and online graphs. By introducing a dustbin row and column, $\mathbf{M}^{(n)}$ is augmented into an assignment score matrix to accommodate unmatched nodes. The resulting assignment scores are normalized using the log-domain Sinkhorn algorithm, producing a soft matching matrix $\mathbf{P}^{(n)}$. The element-wise logarithm of $\mathbf{P}^{(n)}$ is denoted as the matching score matrix $\mathbf{S}^{(n)}=\log \mathbf{P}^{(n)}$.

\paragraph{Matching inference.}
During inference, a distance-constrained mutual nearest-neighbor strategy is adopted to reduce potential mismatches. Specifically, each element $S_{ij}^{(n)}$ in the matching score matrix $\mathbf{S}^{(n)}$ is adjusted according to the spatial distance constraint:

\begin{equation}
\tilde{S}_{i j}^{(n)}=\left\{\begin{array}{lc}
S_{i j}^{(n)}, & \left\|\boldsymbol{p}_{i, \text { off }}-\boldsymbol{p}_{j, \text { on }}\right\|_{2} \leq d_{\text {match }} \\
-\infty, & \text { otherwise }
\end{array} .\right.
\end{equation}

Subsequently, mutual nearest-neighbor\refcite{bib46} filtering is performed on $\tilde{S}_{i j}^{(n)}$, and only candidate matches with matching probabilities no less than the threshold $\tau$ are retained. The matching set at the $n$-th iteration is defined as:

\begin{equation}
\mathcal{M}_{\mathrm{match}}^{(n)}
=
\left\{
(i,j)
\left|
\begin{array}{l}
j=j_i^{*(n)},\ i=i_j^{*(n)},\\
\exp(\tilde{S}_{ij}^{(n)})\geq \tau
\end{array}
\right.
\right\}.
\end{equation}

Here, $j_i^{*(n)}$ denotes the index of the highest-scoring candidate counterpart of the offline node $v_i$ in the online graph, whereas $i_j^{*(n)}$ denotes the index of the highest-scoring candidate counterpart of the online node $v_j$ in the offline graph at the $n$-th iteration. 

\subsection{Confidence estimation}
\paragraph{Matchability confidence estimation.}We further estimate the node matchability confidence from the matching score matrix \(\mathbf{S}^{(n)}\). This confidence reflects the likelihood that a node has a valid correspondence in the other graph, thereby indicating its reliability. It is then used to guide the weight assignment in the subsequent EFA. Specifically, \(\mathbf{S}^{(n)}\) is exponentiated to obtain the soft matching probability matrix \(\mathbf{P}^{(n)}\). The valid matching probability submatrix is extracted by removing the dustbin row and column:

\begin{equation}
\mathbf{P}_{\mathrm{valid}}^{(n)}=
\mathbf{P}_{1:N_{\mathrm{off}},\,1:N_{\mathrm{on}}}^{(n)}
\in \mathbb{R}^{N_{\mathrm{off}}\times N_{\mathrm{on}}}.
\end{equation}

Taking the offline graph as an example, we extract the dustbin-assignment probability vector for offline nodes as
\begin{equation}
\mathbf{p}_{\mathrm{off},\mathrm{bin}}^{(n)}=
\mathbf{P}_{1:N_{\mathrm{off}},\,N_{\mathrm{on}}+1}^{(n)}
\in \mathbb{R}^{N_{\mathrm{off}}}.
\end{equation}

For an offline node \(v_i\), its matchability confidence is estimated by weighting the largest valid matching probability with the probability of not being assigned to the dustbin:

\[c^{(n)}_{i,\mathrm{off}}=
\max_{1 \leq j \leq N_{\mathrm{on}}}
P^{(n)}_{\mathrm{valid}}(i,j)
\left(1-p^{(n)}_{\mathrm{off,bin}}(i)\right).
\]
The first term measures the confidence in the best valid correspondence, while the second term suppresses nodes that are likely to be unmatched. Therefore, the estimated confidence indicates the likelihood that a node has a reliable correspondence.

\paragraph{Matchability confidence estimation with geometric consistency correction.} Although the two-iteration model can produce a relatively stable matching estimate, its confidence estimation can still be affected by ambiguous local structural similarity, interference from non-corresponding nodes, and error accumulation. As a result, some non-corresponding nodes may be assigned overestimated matchability confidence, while the confidence of some truly matchable nodes may be underestimated. To address this issue, CFW-GMN introduces a third iteration to adaptively correct node confidence using geometric consistency constraints. This correction suppresses high-confidence but geometrically inconsistent nodes and retains potentially reliable nodes that better satisfy geometric consistency. To avoid unreliable feedback in the early training stage, a warm-up strategy\refcite{bib47} is adopted, where geometric-consistency confidence feedback is enabled only after several training epochs, when the matching distribution becomes preliminarily stable.
 
Specifically, candidate correspondences with scores higher than the threshold $\tau_{\mathrm{high}}$ are selected from the valid entries of the second-round score matrix $\tilde{S}_{ij}^{(2)}$ to form the high-confidence matching set $\mathcal{M}_{\mathrm{high}}$. RANSAC\refcite{bib48} is then applied to $\mathcal{M}_{\mathrm{high}}$ to robustly estimate a global homography matrix $\mathbf{H}$, which approximates the dominant projective transformation between the two images. Although local geometric distortions may lead to reprojection deviations for some individual nodes, the overall reprojection distribution is expected to remain consistent with the global transformation modeled by $\mathbf{H}$. Therefore, the reprojection error computed under $\mathbf{H}$ is used to quantify geometric consistency.

For an offline node $p_{\mathrm{off}}^i$, its reprojected position in the online image, obtained using $\mathbf{H}$, is denoted as $\hat{p}_{\mathrm{off}}^i$. The nearest-neighbor reprojection errors $d_{\mathrm{off}}^i$ and $d_{\mathrm{on}}^j$ are computed in the online image coordinate system by finding the nearest-neighbor distances between the reprojected offline nodes and the online nodes. To obtain a continuous and numerically stable confidence score, each reprojection error is mapped through a squared-exponential decay function. Specifically, for offline nodes, the geometric consistency confidence is defined as:

\begin{equation}
c^{\mathrm{off}}_{\mathrm{geo},i}=g_{\min}+(1-g_{\min})
\exp\left(-\left(\frac{d^i_{\mathrm{off}}}{s}\right)^2\right),
\end{equation}

where \(s\)  is a scale parameter controlling the decay rate, and \(g_{\min}\) denotes the lower bound imposed on the confidence to avoid numerical instability caused by excessively small confidence values. Based on this, a weighted linear fusion strategy is further adopted to correct the second-round matchability confidence using the geometric consistency confidence. For offline nodes, the corrected confidence is defined as follows:

\[c^{\mathrm{off}}_{\mathrm{corr},i}=
\operatorname{clip}\left((1-\lambda)c^{(2)}_{i,\mathrm{off}}+\lambda c^{\mathrm{off}}_{\mathrm{geo},i}
\right),
\]

where \(\operatorname{clip}(\cdot)\) denotes a clipping function that constrains the corrected confidence within \([0,1]\), and  \(\lambda \in [0,1]\)denotes the fusion weight that controls the contribution of the geometric consistency confidence. The corrected matchability confidence is then fed back to the feature aggregation module to compute the aggregation weights.

\subsection{Loss formulation}
CFW-GMN adopts a two-stage progressive optimization framework. In the early training stage, only the main loss  \(L_{\mathrm{main}}\) is used to optimize the network, ensuring stable convergence of the matching optimization process. After \(e_0\) epochs, the local hard-example mining loss is introduced to improve the model's discriminative ability for ambiguous nodes in local neighborhoods, thereby improving the identification of true correspondences for target nodes.

\paragraph{Main loss.} Two types of supervision labels are defined: ground-truth matching labels and non-corresponding-node labels. Specifically, the set of ground-truth matching index pairs is denoted as $\mathcal{M}_{gt}=\{(i, j)\}$, while the index sets of non-corresponding nodes in the offline and online graphs are denoted as \(\mathcal{L}\) and \(\mathcal{J}\), respectively. Let \(\mathbf{S}^{(2)}=\log \mathbf{P}^{(2)}\), where \(\mathbf{P}^{(2)}\) is the soft matching matrix produced by the second-round Sinkhorn normalization. Based on these labels, a weighted negative log-likelihood loss is constructed on \(\mathbf{S}^{(2)}\). The true matching loss is defined as:

\begin{equation} \ L_{\mathrm{match}} = -\frac{1}{|\mathcal{M}_{gt}|} \sum_{(i,j)\in \mathcal{M}_{gt}} S^{(2)}_{i,j}. \end{equation}

The non-corresponding loss for offline nodes is given by
\begin{equation} \ L^{\mathrm{off}}_{\mathrm{noncorr}} = -\frac{1}{|\mathcal{L}|} \sum_{i\in \mathcal{L}} S^{(2)}_{i, N_{\mathrm{on}}+1}. \end{equation}

Similarly, the non-corresponding loss for online nodes is given by
\begin{equation} \ L ^{\mathrm{on}}_{\mathrm{noncorr}} = -\frac{1}{|\mathcal{J}|} \sum_{j\in \mathcal{J}} S^{(2)}_{N_{\mathrm{off}}+1, j}. \end{equation}

The final main loss combines the true matching loss and the non-corresponding losses:
\[ L_{\mathrm{main}}=
\lambda_{\mathrm{pos}}
\ L_{\mathrm{match}}+
\lambda_{\mathrm{neg}}
\left(
\ L^{\mathrm{off}}_{\mathrm{noncorr}}+
\ L^{\mathrm{on}}_{\mathrm{noncorr}}
\right),
\]

where \(\lambda_{\mathrm{pos}}\) and \(\lambda_{\mathrm{neg}}\) denote the weight coefficients of the true matching term and the non-corresponding term, respectively. Since non-corresponding damage sites are typically more abundant than true correspondences in real-world data, \(\lambda_{\mathrm{pos}}>\lambda_{\mathrm{neg}}\) is adopted to strengthen the supervision of true matches and balance their contributions during optimization. In all experiments, \(\lambda_{\mathrm{pos}}\) and \(\lambda_{\mathrm{neg}}\) are set to \(2.2\) and \(0.8\), respectively.

\begin{figure}[ht]
\centering
\includegraphics[width=0.85\linewidth]{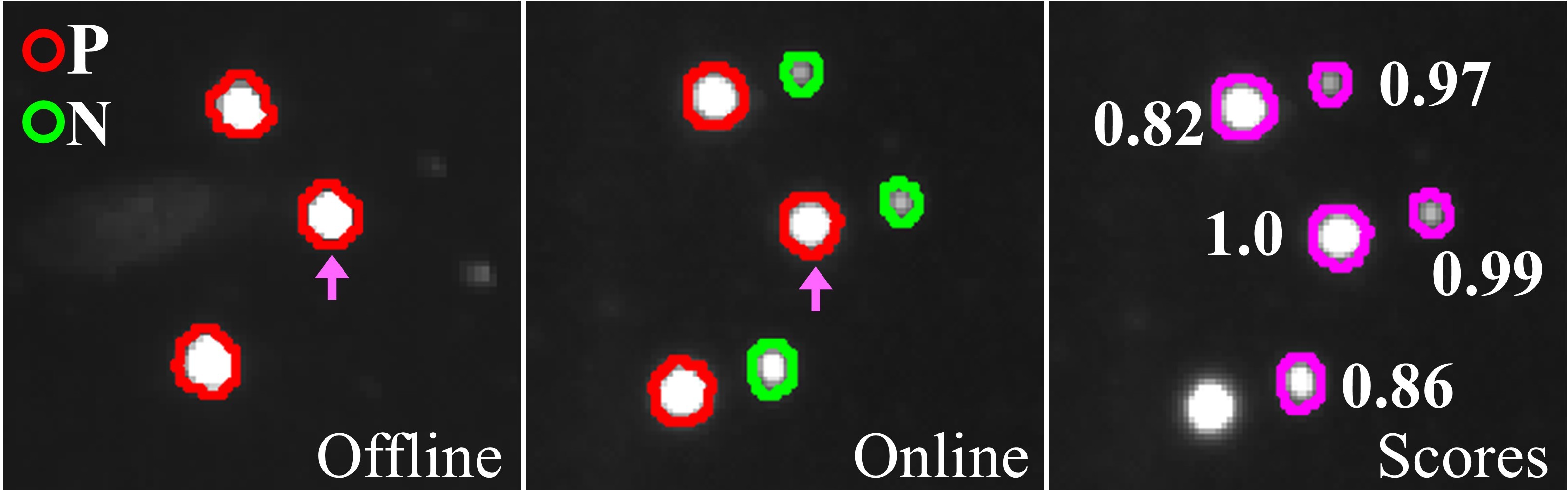}
\caption{Illustration of ambiguity-prone local matching. The matching scores of distractor sites can be close to, or even higher than, those of the true correspondences, thereby causing local matching ambiguity.}
\label{fig:loss}
\end{figure}

\paragraph{Local hard example mining loss.} As shown in \figref{fig:loss}, some distractor sites are located in close proximity to true damage sites and therefore exhibit highly similar geometric features, which can easily lead to local matching ambiguity and mismatches. To further improve the model's ability to distinguish locally confusing samples, we introduce a local hard example mining loss, \(L_{\mathrm{hard}}\), in addition to the primary matching loss, \(L_{\mathrm{main}}\). This loss imposes a margin constraint between each true correspondence and its nearby hard samples, thereby enlarging the score separation between positive and negative matches, reducing local ambiguity, and improving matching accuracy.

Taking the offline branch as an example, \(\mathcal{M}_{\mathrm{off}}\) denotes the set of ground-truth damaged-site indices with local hard samples. For each \(i \in \mathcal{M}_{\mathrm{off}}\), nearby interfering sites within radius \(d_{\mathrm{hard}}\) are collected as \(\mathcal{N}^{\mathrm{off}}_{d_{\mathrm{hard}}}(i)\). Using the score matrix \(\mathbf{S}^{(3)}\) produced by the third-round optimal transport module, the local hard-sample constraint requires the ground-truth matching score \(S_{i,j}^{(3)}\) to exceed each local hard-sample score \(S_{\mu,j}^{(3)}\) for \(\mu \in \mathcal{N}^{\mathrm{off}}_{d_{\mathrm{hard}}}(i)\) by at least a predefined margin \(m\). The loss is defined as:

\begin{equation}
\begin{aligned}
L_{\mathrm{hard}}^{\mathrm{off}}
&=
\frac{1}{\lvert\mathcal{M}_{\mathrm{off}}\rvert}
\sum_{i\in\mathcal{M}_{\mathrm{off}}}
\frac{1}{\mathcal{H}_{i}^{\mathrm{off}}}
\\
&\quad \times
\sum_{\mu\in\mathcal{N}^{\mathrm{off}}_{d_{\mathrm{hard}}}(i)}
\left[
m - S_{i,j}^{(3)} + S_{\mu,j}^{(3)}
\right]_{+},
\end{aligned}
\end{equation}

where \([x]_+=\max(x,0)\) denotes the hinge function, and \(\mathcal{H}_{i}^{\mathrm{off}}\) is the number of local hard samples yielding non-zero hinge loss values for site (i). If \(\mathcal{H}_{i}^{\mathrm{off}}=0\), the corresponding hard-sample loss term is set to zero. Finally, the total loss is defined by combining the offline and online hard-sample losses as:

\begin{equation}
L=\begin{cases}
L_{\mathrm{main}}, & e \leq e_{0}, \\[2mm]
L_{\mathrm{main}}+\lambda_{\mathrm{hard}}\left(L_{\mathrm{hard}}^{\mathrm{off}}+L_{\mathrm{hard}}^{\mathrm{on}}\right), & e > e_{0}.
\end{cases}
\end{equation}

Here, \(e\) denotes the current training epoch, \(e_0\) is the warm-up threshold, and \(\lambda_{\mathrm{hard}}\) controls the contribution of the local hard-sample loss.

\subsection{Matching post-processing (MPP)}
The predicted correspondences produced by the network are used to estimate the homography matrix \(\mathbf{H}\). Geometric consistency verification is then performed based on the reprojection error to remove erroneous matches that deviate significantly from the estimated geometric constraint, while retaining candidate matches that satisfy geometric consistency. The retained correspondences are used to form the final matching set \(\mathcal{M}_{\mathrm{final}}\). Since this procedure follows a similar strategy to that described in Section 2.4, its details are omitted here. It should be noted that, due to geometric distortions, a single estimated global homography cannot perfectly account for all correspondences; rather, it serves as an approximate geometric constraint for match refinement. The effectiveness of MPP depends on the reliability of the initial correspondences predicted by the network. When the estimated homography is sufficiently reliable, geometric verification can further refine the matching set and improve the overall matching performance.

\section{Experiments}
\subsection{FODI damage sites matching dataset}
Online full-aperture images of optics in the CAEP laser facility were acquired using the FODI system. After removal, purging, and cleaning, the same optics were inspected on an offline platform to obtain the corresponding offline full-aperture images. A real-data damage-site matching dataset was constructed using our previously developed method\refcite{bib49}, followed by manual verification. Constrained by facility operation and maintenance conditions and annotation costs, the acquisition and high-quality annotation of real data are challenging, resulting in a relatively limited dataset size.

\begin{figure*}[!b]
\centering
\includegraphics[width=0.95\textwidth]{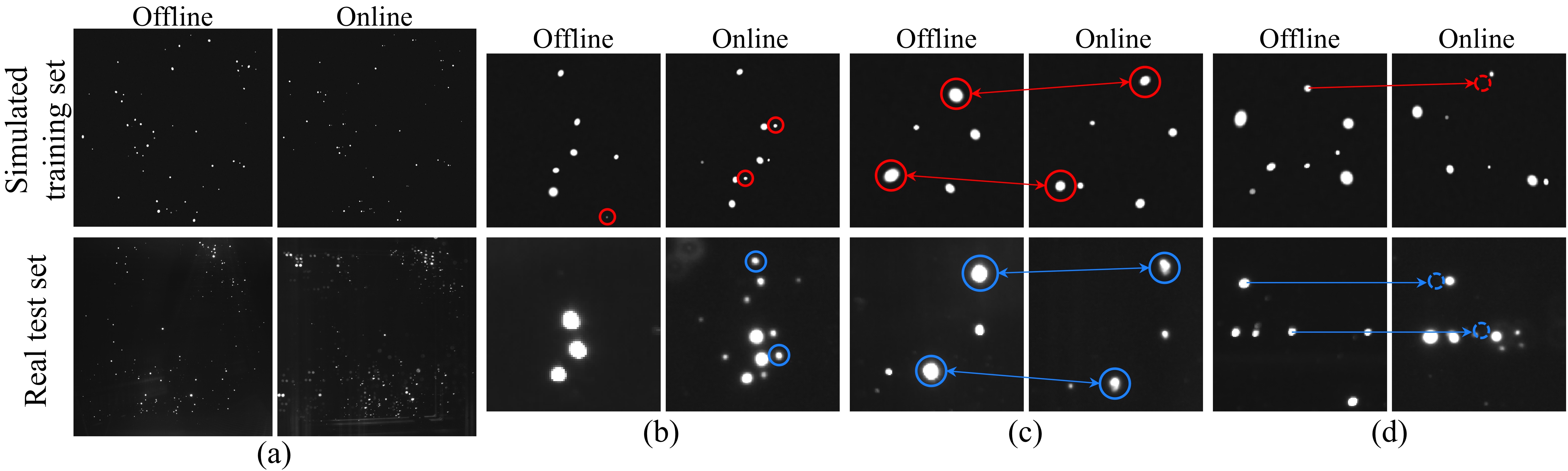}
\caption{Representative samples from the simulated training set and the real-data test set. (a) Spatially Non-uniform Damage Distribution. (b) Interference from Non-corresponding Damage Sites. (c) Morphological Deformation of Damage Sites. (d) Local positional offset.}
\label{fig:train}
\end{figure*}

\paragraph{Simulated training dataset.} To provide sufficient training samples and account for diverse variations in real scenarios, we constructed a synthetic training dataset. In the simulation, the intensity profile of each damage site was modeled using a point spread function (PSF) parameterized by a super-Gaussian kernel. To improve sample realism, spatially non-uniform damage distributions, morphological deformations, local positional offsets, and non-corresponding damage sites were further introduced. \figref{fig:train} shows representative simulated damage samples. The training set contains 200 image pairs, including 11,444 offline damage sites and 16,257 online damage sites. The proportions of matched damage sites on the offline and online sides are approximately 75\% and 53\%, respectively.

\paragraph{Validation set.} The validation set was used only for model selection, early stopping, and hyperparameter tuning. It consists of 8 pairs of real samples, containing 2946 offline damage sites and 2528 online damage sites. The proportions of matched damage sites on the offline and online sides are approximately 31.70$\%$ and 36.95$\%$ respectively.

\paragraph{Test Set.} To evaluate the proposed method under different levels of interference, we constructed two independent test sets, Simplified-Scene and Complex-Scene, both separated from the validation set to avoid information leakage. Simplified-Scene removes part of the non-corresponding damage sites to reduce interference, containing 627 offline and 778 online damage sites from 11 real sample pairs, with matched-site ratios of 77.51$\%$ and 62.47$\%$, respectively. Complex-Scene uses the unfiltered raw data from the same 11 sample pairs, containing 1,275 offline and 2,083 online damage sites, with matched-site ratios of 71.92$\%$ and 44.02$\%$, respectively.

Although the number of real image pairs is relatively limited, each pair contains sufficient damage sites and representative interference factors. Thus, the test sets can reasonably reflect the matching performance of the model in practical applications.

\paragraph{Experimental settings.} CFW-GMN was trained end-to-end using the Adam optimizer with an initial learning rate of \(1\times10^{-4}\). A StepLR learning-rate decay schedule was adopted, where the learning rate was multiplied by 0.5 every 5 epochs. The batch size was set to 20, and the maximum number of training epochs was set to 60. The model parameters that achieved the highest validation accuracy were selected as the final model. To improve training stability, a two-stage training strategy was employed, where the training entered the second stage once the validation accuracy plateaued. 

\subsection{Experimental results}
\begin{figure*}[!t]
\centering
\includegraphics[width=0.75\textwidth]{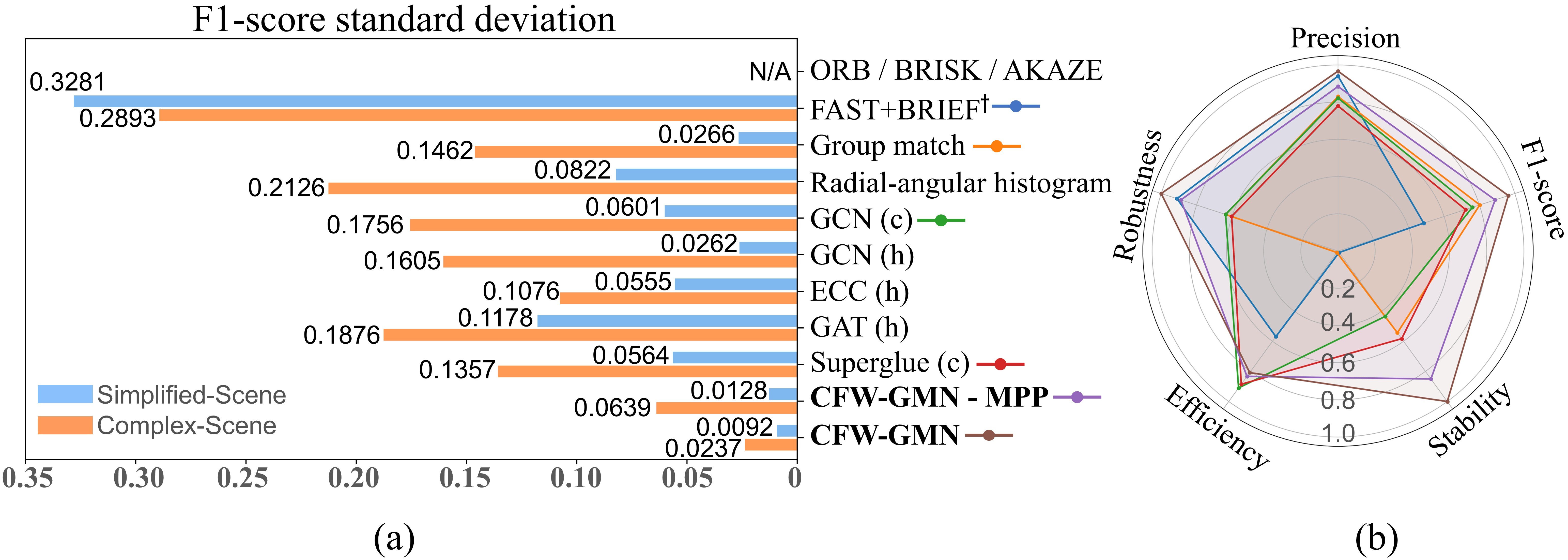}
\caption{Performance comparison of different matching methods. (a) Bar chart of the F1-score standard deviation across different samples. (b) Radar chart of multi-metric matching performance.}
\label{fig:compare}
\end{figure*}

\begin{figure*}[htbp]
\centering
\includegraphics[width=0.98\textwidth]{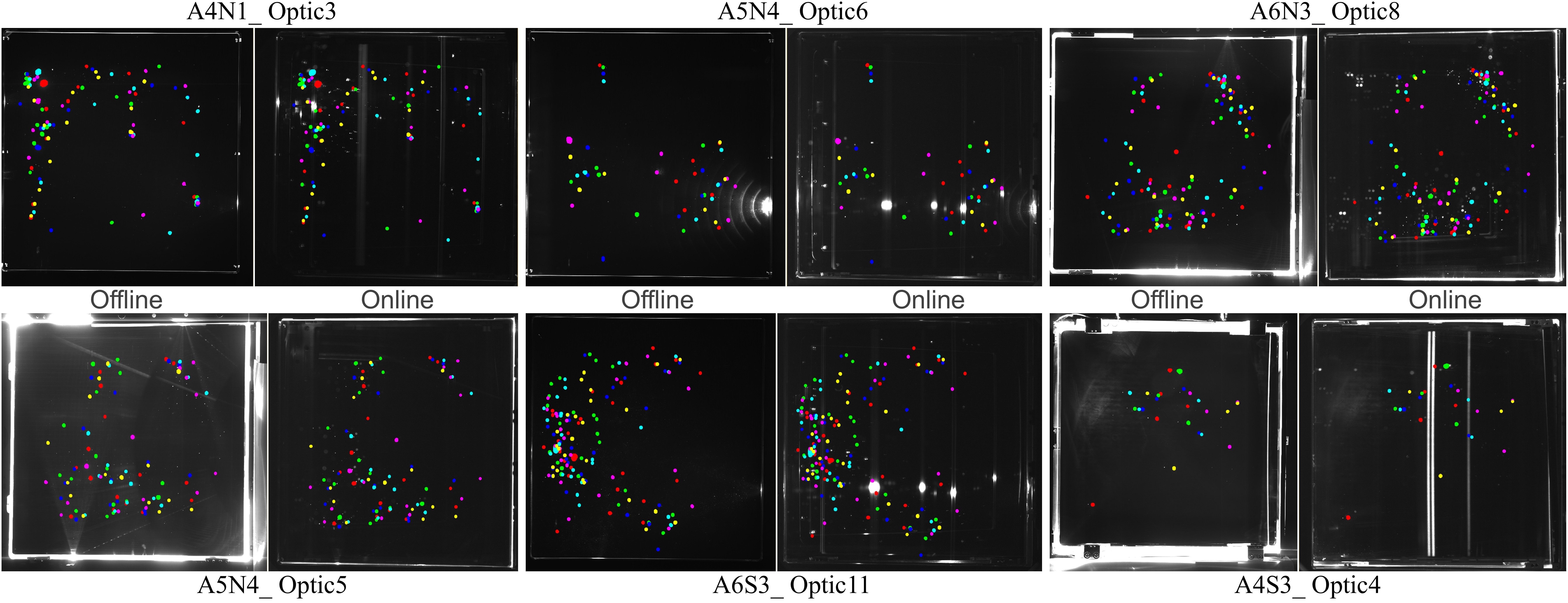}
\caption{Matching results of CFW-GMN on the Complex-Scene dataset. Damage sites with the same color at corresponding online-offline locations indicate predicted correspondences, while repeated colors may appear due to the limited color palette.}
\label{fig:final}
\end{figure*}

We compared the proposed method with 11 baseline algorithms from three categories on the Simplified-Scene and Complex-Scene test sets. The quantitative matching results are reported in ~\tabref{tab:matching_results} and \figref{fig:compare}, and the visualized matching results on Complex-Scene are shown in \figref{fig:final}. To enable traditional local-descriptor-based methods to produce effective matches, each original image was uniformly partitioned into four subregions, and matching was performed independently within each subregion. The reported results for these methods were obtained under this setting. In addition, each GNN-based baseline was evaluated using either coordinates or radial-angular histograms as input features, and the better result is reported to avoid underestimating its performance.

\begin{table*}[htbp]
\centering
\caption{Comparison of matching performance on the Simplified-Scene and Complex-Scene datasets.}
\label{tab:matching_results}
\footnotesize
\setlength{\tabcolsep}{1.8pt}
\renewcommand{\arraystretch}{1.12}
\begin{threeparttable}
\begin{tabular}{@{}ll*{5}{c}*{5}{c}c@{}}
\toprule
\multirow{2}{*}{Category} 
& \multirow{2}{*}{Matching method} 
& \multicolumn{5}{c}{Simplified-Scene} 
& \multicolumn{5}{c}{Complex-Scene} 
& \multirow{2}{*}{\makecell{Performance drop\\F1-score (\%)}} \\
\cmidrule(lr){3-7} \cmidrule(lr){8-12}
& & Accuracy & Precision & Recall & F1-score & Time (s)
  & Accuracy & Precision & Recall & F1-score & Time (s) & \\
\midrule
\multirow{2}{*}{\makecell{Local\\descriptors}}
& ORB / BRISK / AKAZE
& N/A & N/A & N/A & N/A & N/A
& N/A & N/A & N/A & N/A & N/A
& N/A \\
& FAST+BRIEF$^{\dagger}$
& 58.45 & 94.05 & 41.67 & 51.79 & 17.45
& 67.93 & 94.05 & 37.28 & 48.49 & 17.48
& 6.37 \\
\midrule
\multirow{2}{*}{\makecell{Star\\identification}}
& Group Match
& 95.31 & 94.36 & 99.36 & 96.77 & 12.19
& 81.10 & 83.02 & 81.75 & 80.20 & 39.69
& 17.12 \\
& Radial-Angular Histogram
& 80.95 & 95.71 & 75.75 & 84.02 & 0.34
& 67.04 & 80.29 & 46.42 & 56.92 & 0.35
& 32.25 \\
\midrule
\multirow{5}{*}{GNN}
& GCN (coordinate)
& 89.09 & 92.10 & 91.46 & 91.68 & 3.25
& 77.93 & 82.21 & 72.58 & 76.00 & 3.94
& 17.10 \\
& GCN (histogram)
& 94.25 & 97.04 & 94.28 & 95.57 & 3.28
& 81.26 & 92.06 & 68.92 & 76.06 & 3.94
& 20.41 \\
& ECC (histogram)
& 83.83 & 92.60 & 82.75 & 87.18 & 3.73
& 75.23 & 86.54 & 63.10 & 71.77 & 4.95
& 17.68 \\
& GAT (histogram)
& 63.38 & 89.88 & 52.19 & 65.12 & 3.87
& 57.64 & 78.27 & 26.99 & 38.06 & 4.96
& 41.55 \\
& SuperGlue (coordinate)
& 84.19 & 89.98 & 87.28 & 88.44 & 4.16
& 73.73 & 78.80 & 67.84 & 72.12 & 4.91
& 18.45 \\
\midrule
\multirow{2}{*}{\textbf{Ours}}
& \textbf{CFW-GMN - MPP}
& 94.15 & 94.82 & 96.80 & \textbf{95.78} & 5.10
& 88.51 & 88.45 & 89.67 & \textbf{88.85} & 6.99
& \textbf{7.24} \\
& \textbf{CFW-GMN}
& 98.94 & 98.49 & 100.00 & \textbf{99.23} & 5.65
& 96.33 & 96.70 & 96.18 & \textbf{96.36} & 7.97
& \textbf{2.89} \\
\bottomrule
\end{tabular}
\begin{tablenotes}[flushleft]
\footnotesize
\item Note: Accuracy, Precision, Recall, and F1-score are reported in percentages. N/A indicates that the method could not produce valid results under the unified evaluation protocol, e.g., due to insufficient valid correspondences or failed RANSAC-based geometric estimation. $^{\dagger}$ indicates that FAST+BRIEF achieves a success rate of 0.73, and the reported metrics are computed only on successfully matched samples. All other methods with valid results achieve a success rate of 1.00.
\end{tablenotes}
\end{threeparttable}
\end{table*}

The experimental results show that traditional local-descriptor-based matching methods exhibit limited overall performance on large-scale weak-feature image matching tasks. Even in the Simplified-Scene setting with relatively few interfering points, these methods still suffer from insufficient valid correspondences or matching failures. In contrast, the star-identification-based Group Match method performs well on the Simplified-Scene dataset, achieving an F1-score of 96.77$\%$. However, when the number of interfering sites increases, its F1-score decreases by 16.57 percentage points, and its runtime reaches 39.69 s, making it less suitable for rapid inspection scenarios. Although the Radial-Angular Histogram method achieves the highest computational efficiency, with a runtime of approximately 0.35 s, its overall matching performance remains limited. Compared with the above two categories of methods, GNN-based methods improve both matching performance and computational efficiency. Nevertheless, under weak-feature inputs and complex interference, their feature representation and matching discrimination capabilities remain insufficient, as reflected by F1-scores remaining around 70$\%$.

Overall, the proposed CFW-GMN achieves F1-scores of 99.23$\%$ and 96.36$\%$ on the Simplified-Scene and Complex-Scene test sets, respectively, outperforming all baseline methods. It also yields the lowest F1-score standard deviation and performance degradation, with values of only 0.0237 and 2.89$\%$, respectively, indicating strong robustness and stability under complex matching conditions. The average processing time is approximately 0.7 s per image pair, making the method suitable for fast matching scenarios. As shown in the radar chart, CFW-GMN achieves balanced performance across multiple metrics, demonstrating its overall effectiveness and adaptability to complex scenes. Moreover, even without the post-processing module, CFW-GMN still maintains an end-to-end F1-score close to 90$\%$ on the Complex-Scene dataset.

\section{Ablation Study}
We sequentially remove the core components of the proposed pipeline in reverse order. The removed components include the local hard example mining loss, the third-round iterative module including EFA with geometry-corrected matchability confidence weighting, RFF positional encoding, local self- and cross-attention, and the second-round iterative module including EFA with matchability confidence weighting. The resulting variants are evaluated to analyze the contribution of each module to damage-site matching and its effect on geometric feature aggregation. The results are reported in ~\tabref{tab:ablation}.

\begin{table}[t]
\centering
\caption{Ablation study results on the Complex-Scene dataset.}
\label{tab:ablation}
\small
\renewcommand{\arraystretch}{1.12}

\begin{tabular*}{\linewidth}{@{\extracolsep{\fill}}lcccc@{}}
\toprule
Configuration & Accuracy & Precision & Recall & F1-score \\
\midrule
CFW-GMN & 96.33 & 96.70 & 96.18 & 96.36 \\
$-$ MPP & 88.51 & 88.45 & 89.67 & 88.85 \\
$-$ Hard loss & 86.57 & 87.61 & 86.10 & 86.61 \\
$-$ Third iteration & 81.78 & 84.18 & 78.67 & 80.95 \\
$-$ Second iteration & 76.45 & 81.23 & 69.72 & 74.24 \\
$-$ RFF encoding & 68.14 & 84.96 & 46.96 & 57.70 \\
$-$ Local attention & 65.56 & 92.00 & 39.24 & 50.36 \\
\bottomrule
\end{tabular*}

\vspace{2pt}
\begin{minipage}{0.98\linewidth}
\footnotesize
\textit{Note:} Accuracy, Precision, Recall, and F1-score are reported in percentage.
\end{minipage}
\end{table}

The progressive removal of key components leads to varying degrees of performance degradation, indicating that the designed network structure and functional components contribute positively to the final matching performance. Specifically, when MPP and the local hard example mining loss are removed, CFW-GMN exhibits a performance drop but remains relatively high, suggesting that these two components enhance the discriminative capability of the model to some extent. In contrast, removing the third and second iterative modules results in a significant decrease in the F1 score, demonstrating that confidence-feedback-weighted EFA is essential for constructing effective geometric relationship representations. Furthermore, removing RFF positional encoding and local attention leads to an approximately 20-percentage-point decrease in recall. This suggests that spatial geometric relationship representation and local feature interaction serve as important foundations for the model to reliably capture true correspondences. Overall, these results demonstrate that the proposed modules effectively improve the matching performance, robustness to interference, and stability of the model.

\begin{figure*}[htbp]
\centering
\includegraphics[width=0.98\textwidth]{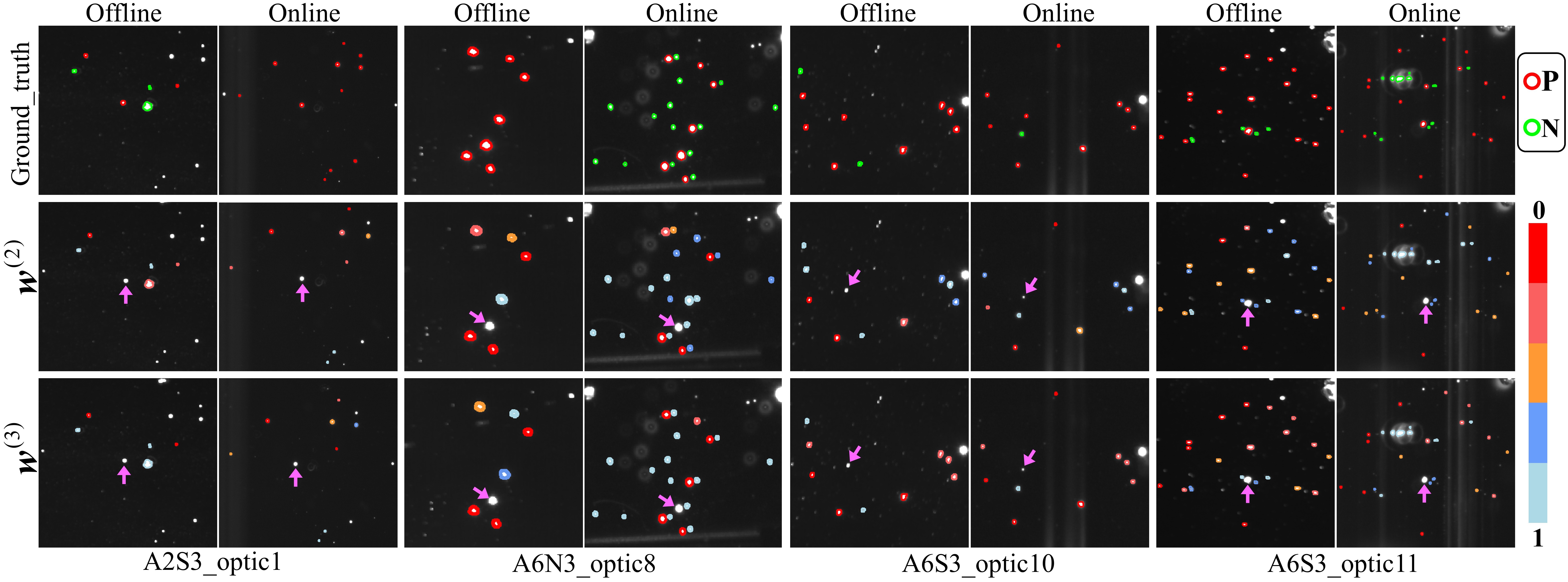}
\caption{Visualization of edge-feature weights during feature aggregation in Stages 2 and 3. Red regions indicate matched true damage sites. Green regions indicate non-corresponding damage sites. }
\label{fig:weight-show}
\end{figure*}

\figref{fig:weight-show} visualizes the edge feature weights during feature aggregation in Stages 2 and 3. Compared with the ground truth, the learned weights indicate that, as the iterations proceed, the edge weights between non-corresponding nodes gradually decrease, whereas truly corresponding nodes are assigned higher aggregation weights and progressively become the main information sources for node feature updates. This suggests that the proposed confidence-feedback iterative aggregation mechanism can use matching inference information to guide the model in assessing the reliability of neighboring relationships. In this way, the propagation of interfering information can be suppressed, while the role of reliable neighboring relationships in feature aggregation is enhanced. This mechanism provides the model with a selective suppression capability, allowing the attention mechanism to exploit reliable neighborhood information more effectively.

\section{Conclusion}
To address the challenging problem of matching images of weak-feature laser-induced damage sites on large-aperture optics in high-power laser facilities, this paper proposes a confidence-feedback-weighted graph matching network. The proposed method enables high-precision damage site matching under complex conditions. Experimental results on the Simplified-Scene and Complex-Scene datasets show that, even when the proportion of interfering sites reaches 55.98$\%$, the proposed method can suppress the influence of interfering sites during feature interaction through the confidence-feedback iterative aggregation mechanism, enhance reliable node information, and construct more discriminative cross-graph feature representations. As a result, reliable matching site pairs are accurately identified, achieving an F1-score of 96.36$\%$. Compared with existing methods, CFW-GMN achieves a substantial improvement in matching accuracy while demonstrating good robustness and stability. Its F1-score standard deviation across different samples is only 0.0237, and its F1-score decreases by only 2.89 percentage points from the Simplified-Scene dataset to the Complex-Scene dataset. Moreover, while maintaining high matching accuracy, the proposed method requires only 0.7248 s per image pair on average, satisfying the requirement for rapid matching in practical ICF experiments. Future work will further investigate more discriminative feature construction and representation methods to improve damage site matching performance under complex and challenging conditions.

%\appendix{} 附录


\begin{thebibliography}{00}

\bibitem{bib1} C. Osolin, ``Target Breakthrough Enabled Fusion Record at NIF'', https://lasers.llnl.gov/news/target-breakthrough-enabled-fusion-record-nif (April 15, 2026). 

\bibitem{bib2} Z. Liu, J. Zhang, S. Wang, F. Geng, Q. Zhang, J. Cheng, M. Chen, and Q. Xu, ``Ultrafast Process Characterization of Laser-Induced Damage in Fused Silica Using Pump-Probe Shadow Imaging Techniques'', Materials 17, 837 (2024). DOI: https://doi.org/10.3390/ma17040837

\bibitem{bib3} C. Lacombe, L. Lamaignère, G. Hallo, M. Sozet, T. Donval, G. Raz{'e}, C. Ameil, M. Benoit, F. Gaudfrin, E. Bordenave, N. Bonod, and J. N{'e}auport, ``Full-scale optic designed for onsite study of damage growth at the Laser MegaJoule facility'', Opt. Express 31, 4291--4305 (2023). DOI: https://doi.org/10.1364/OE.474581

\bibitem{bib4} Z. M. Liao, M. Nostrand, P. Whitman, and J. Bude, ``Analysis of optics damage growth at the National Ignition Facility'', Proc. SPIE 9632, 963217 (2015). DOI: https://doi.org/10.1117/12.2195515

\bibitem{bib5} X. Hu, W. Zhou, H. Guo, X. Huang, B. Zhao, W. Zhong, Q. Zhu, and Z. Chen, ``The Prediction of Incremental Damage on Optics from the Final Optic Assembly in an ICF High-Power Laser Facility'', Appl. Sci. 14, 5226 (2024). DOI: https://doi.org/10.3390/app14125226

\bibitem{bib6} A. D. Conder, J. J. Chang, L. M. Kegelmeyer, M. L. Spaeth, and P. K. Whitman, ``Final Optics Damage Inspection (FODI) for the National Ignition Facility'', Proc. SPIE 7797, 77970P (2010). DOI: https://doi.org/10.1117/12.862596

\bibitem{bib7} L. M. Kegelmeyer, R. Clark, R. R. Leach, D. McGuigan, V. M. Kamm, D. Potter, J. T. Salmon, J. Senecal, A. Conder, M. Nostrand, and P. K. Whitman, ``Automated optics inspection analysis for NIF'', Fusion Eng. Des. 87, 2120--2124 (2012). DOI: https://doi.org/10.1016/j.fusengdes.2012.09.017

\bibitem{bib8} F. P. Wei, F. D. Chen, J. Tang, Z. T. Peng, and G. D. Liu, ``Final optics damage online inspection in high power laser facility'', Optoelectron. Lett. 15, 306--311 (2019). DOI: https://doi.org/10.1007/s11801-019-8193-3

\bibitem{bib9} F. Wei, F. Chen, B. Liu, Z. Peng, J. Tang, Q. Zhu, D. Hu, Y. Xiang, N. Liu, Z. Sun, and G. Liu, ``Automatic classification of true and false laser-induced damage in large aperture optics'', Opt. Eng. 57, 053112 (2018). DOI: https://doi.org/10.1117/1.OE.57.5.053112
\bibitem{bib10} Y. Li, J. L. Li, Z. Li, D. A. Liu, D. W. Zhang, and J. Y. Zhang, ``Inspection and Repair of Optical Damage in Tradition and Deep Learning (Invited)'', Acta Photonica Sin. 51, 1012002 (2022). DOI: https://doi.org/10.3788/gzxb20225110.1012002

\bibitem{bib11} W. Zheng, X. Wei, Q. Zhu, F. Jing, D. Hu, J. Su, K. Zheng, X. Yuan, H. Zhou, W. Dai, W. Zhou, F. Wang, D. Xu, X. Xie, B. Feng, Z. Peng, L. Guo, Y. Chen, X. Zhang, L. Liu, D. Lin, Z. Dang, Y. Xiang, and X. Deng, ``Laser performance of the SG-III laser facility'', High Power Laser Sci. Eng. 4, e21 (2016). DOI: https://doi.org/10.1017/hpl.2016.20

\bibitem{bib12} J. E. Heebner, ``Injection laser system architecture, upgrades, and future at the National Ignition Facility'', Proc. SPIE 13358, 133580F (2025). DOI: https://doi.org/10.1117/12.3040268

\bibitem{bib13} C. W. Carr, ``Fusion enabling laser-induced damage reduction, management, and repair strategies at the National Ignition Facility'', Proc. SPIE 12726, 1272602 (2023). DOI: https://doi.org/10.1117/12.2688417

\bibitem{bib14} X. Chai, P. Li, J. Zhao, G. Wang, D. Zhu, Y. Jiang, B. Chen, Q. Zhu, B. Feng, L. Wang, and Y. Jing, ``Laser-induced damage growth of large-aperture fused silica optics under high-fluence 351 nm laser irradiation'', Optik 226, 165549 (2021). DOI: https://doi.org/10.1016/j.ijleo.2020.165549

\bibitem{bib15} X. Liang, J. Sun, X. Wang, J. Li, L. Zhang, and J. Guo, ``Surface weak scratch detection for optical elements based on a multimodal imaging system and a deep encoder--decoder network'', J. Opt. Soc. Am. A 40, 1237--1248 (2023). DOI: https://doi.org/10.1364/JOSAA.483381

\bibitem{bib16} M. L. Spaeth, P. J. Wegner, T. I. Suratwala, M. C. Nostrand, J. D. Bude, A. D. Conder, J. A. Folta, J. E. Heebner, L. M. Kegelmeyer, B. J. MacGowan, D. C. Mason, M. J. Matthews, and P. K. Whitman, ``Optics Recycle Loop Strategy for NIF Operations Above UV Laser-Induced Damage Threshold'', Fusion Sci. Technol. 69, 265--294 (2016). DOI: https://doi.org/10.13182/FST15-119

\bibitem{bib17}
D. G. Lowe, ``Distinctive image features from scale-invariant keypoints'', Int. J. Comput. Vis. 60, 91--110 (2004). DOI: https://doi.org/10.1023/B:VISI.0000029664.99615.94
\bibitem{bib18}
H. Bay, T. Tuytelaars, and L. Van Gool, ``SURF: Speeded Up Robust Features'', in \emph{Computer Vision -- ECCV 2006}, Lecture Notes Comput. Sci. 3951, 404--417 (2006). DOI: https://doi.org/10.1007/11744023\_32
\bibitem{bib19}
E. Rosten, R. Porter, and T. Drummond, ``Faster and better: A machine learning approach to corner detection'', IEEE Trans. Pattern Anal. Mach. Intell. 32, 105--119 (2010). DOI: https://doi.org/10.1109/TPAMI.2008.275
\bibitem{bib20}
M. Calonder, V. Lepetit, M. Özuysal, T. Trzcinski, C. Strecha, and P. Fua, ``BRIEF: Computing a Local Binary Descriptor Very Fast'', IEEE Trans. Pattern Anal. Mach. Intell. 34, 1281--1298 (2012). DOI: https://doi.org/10.1109/TPAMI.2011.222
\bibitem{bib21}
E. Rublee, V. Rabaud, K. Konolige, and G. Bradski, ``ORB: An efficient alternative to SIFT or SURF'', in \emph{Proc. IEEE Int. Conf. Comput. Vis.} 2564--2571 (2011). DOI: https://doi.org/10.1109/ICCV.2011.6126544
\bibitem{bib22}
S. Leutenegger, M. Chli, and R. Y. Siegwart, ``BRISK: Binary Robust Invariant Scalable Keypoints'', in \emph{Proc. IEEE Int. Conf. Comput. Vis.} 2548--2555 (2011). DOI: https://doi.org/10.1109/ICCV.2011.6126542
\bibitem{bib23}
P. F. Alcantarilla, J. Nuevo, and A. Bartoli, ``Fast Explicit Diffusion for Accelerated Features in Nonlinear Scale Spaces'', in \emph{Proc. Br. Mach. Vis. Conf.} 13.1--13.11 (2013). DOI: https://doi.org/10.5244/C.27.13
\bibitem{bib24}
D. Rijlaarsdam, H. Yous, J. Byrne, D. Oddenino, G. Furano, and D. Moloney, ``A Survey of Lost-in-Space Star Identification Algorithms Since 2009'', Sensors 20, 2579 (2020). DOI: https://doi.org/10.3390/s20092579
\bibitem{bib25}
B. B. Spratling IV and D. Mortari, ``A Survey on Star Identification Algorithms'', Algorithms 2, 93--107 (2009). DOI: https://doi.org/10.3390/a2010093
\bibitem{bib26}
D. Mortari, M. A. Samaan, C. Bruccoleri, and J. L. Junkins, ``The Pyramid Star Identification Technique'', Navigation 51, 171--183 (2004). DOI: https://doi.org/10.1002/j.2161-4296.2004.tb00349.x
\bibitem{bib27}
M. Kolomenkin, S. Pollak, I. Shimshoni, and M. Lindenbaum, ``Geometric voting algorithm for star trackers'', IEEE Trans. Aerosp. Electron. Syst. 44, 441--456 (2008). DOI: https://doi.org/10.1109/TAES.2008.4560198
\bibitem{bib28}
H. Lee and H. Bang, ``Star pattern identification technique by modified grid algorithm'', IEEE Trans. Aerosp. Electron. Syst. 43, 1112--1116 (2007). DOI: https://doi.org/10.1109/TAES.2007.4383600
\bibitem{bib29}
G. Zhang, X. Wei, and J. Jiang, ``Full-sky autonomous star identification based on radial and cyclic features of star pattern'', Image Vis. Comput. 26, 891--897 (2008). DOI: https://doi.org/10.1016/j.imavis.2007.10.006
\bibitem{bib30}
J. Fu, L. Lin, and Q. Li, ``Spherical Polar Pattern Matching for Star Identification'', Sensors 25, 4201 (2025). DOI: https://doi.org/10.3390/s25134201

\bibitem{bib31}
F. Scarselli, M. Gori, A. C. Tsoi, M. Hagenbuchner, and G. Monfardini, ``The Graph Neural Network Model'', IEEE Trans. Neural Netw. 20, 61--80 (2009). DOI: https://doi.org/10.1109/TNN.2008.2005605
\bibitem{bib32}
Y. Li, C. Gu, T. Dullien, O. Vinyals, and P. Kohli, ``Graph Matching Networks for Learning the Similarity of Graph Structured Objects'', in \emph{Proc. 36th Int. Conf. Mach. Learn.}, Proc. Mach. Learn. Res. 97, 3835--3845 (2019). https://proceedings.mlr.press/v97/li19d.html
\bibitem{bib33}
J. Gilmer, S. S. Schoenholz, P. F. Riley, O. Vinyals, and G. E. Dahl, ``Neural Message Passing for Quantum Chemistry'', in \emph{Proc. 34th Int. Conf. Mach. Learn.}, Proc. Mach. Learn. Res. 70, 1263--1272 (2017). https://proceedings.mlr.press/v70/gilmer17a.html
\bibitem{bib34}
T. N. Kipf and M. Welling, ``Semi-Supervised Classification with Graph Convolutional Networks'', in \emph{Int. Conf. Learn. Represent.} (2017). https://openreview.net/forum?id=SJU4ayYgl
\bibitem{bib35}
M. Simonovsky and N. Komodakis, ``Dynamic Edge-Conditioned Filters in Convolutional Neural Networks on Graphs'', in \emph{Proc. IEEE Conf. Comput. Vis. Pattern Recognit.} 3693--3702 (2017). DOI: https://doi.org/10.1109/CVPR.2017.11
\bibitem{bib36}
P. Veličković, G. Cucurull, A. Casanova, A. Romero, P. Liò, and Y. Bengio, ``Graph Attention Networks'', in \emph{Int. Conf. Learn. Represent.} (2018). https://openreview.net/forum?id=rJXMpikCZ
\bibitem{bib37}
P.-E. Sarlin, D. DeTone, T. Malisiewicz, and A. Rabinovich, ``SuperGlue: Learning Feature Matching with Graph Neural Networks'', in \emph{Proc. IEEE/CVF Conf. Comput. Vis. Pattern Recognit.} 4938--4947 (2020). DOI: https://doi.org/10.1109/CVPR42600.2020.00499
\bibitem{bib38}
J. Sun, Z. Shen, Y. Wang, H. Bao, and X. Zhou, ``LoFTR: Detector-Free Local Feature Matching with Transformers'', in \emph{Proc. IEEE/CVF Conf. Comput. Vis. Pattern Recognit.} 8922--8931 (2021). DOI: https://doi.org/10.1109/CVPR46437.2021.00881

\bibitem{bib39}
M.-T. Luong, H. Pham, and C. D. Manning, ``Effective Approaches to Attention-based Neural Machine Translation'', in \emph{Proc. 2015 Conf. Empirical Methods Nat. Lang. Process.} 1412--1421 (2015). DOI: https://doi.org/10.18653/v1/D15-1166
\bibitem{bib40}
A. Vaswani, N. Shazeer, N. Parmar, J. Uszkoreit, L. Jones, A. N. Gomez, L. Kaiser, and I. Polosukhin, ``Attention Is All You Need'', in \emph{Adv. Neural Inf. Process. Syst.} 30, 5998--6008 (2017). https://papers.nips.cc/paper/7181-attention-is-all-you-need

\bibitem{bib41}
Y. Han, Y. Huang, H. Dong, F. Chen, F. Zeng, Z. Peng, Q. Zhu, and G. Liu, ``Continuous gradient fusion class activation mapping: segmentation of laser-induced damage on large-aperture optics in dark-field images'', High Power Laser Sci. Eng. 12, e4 (2024). DOI: https://doi.org/10.1017/hpl.2023.85
\bibitem{bib42}
A. Rahimi and B. Recht, ``Random Features for Large-Scale Kernel Machines'', in \emph{Adv. Neural Inf. Process. Syst.} 20 (2007). https://papers.nips.cc/paper/3182-random-features-for-large-scale-kernel-machines
\bibitem{bib43}
M. Tancik, P. P. Srinivasan, B. Mildenhall, S. Fridovich-Keil, N. Raghavan, U. Singhal, R. Ramamoorthi, J. T. Barron, and R. Ng, ``Fourier Features Let Networks Learn High Frequency Functions in Low Dimensional Domains'', in \emph{Adv. Neural Inf. Process. Syst.} 33 (2020). arXiv: https://arxiv.org/abs/2006.10739


\bibitem{bib44}
M. Cuturi, ``Sinkhorn Distances: Lightspeed Computation of Optimal Transport'', in \emph{Adv. Neural Inf. Process. Syst.} 26, 2292--2300 (2013). https://papers.nips.cc/paper/4927-sinkhorn-distances-lightspeed-computation-of-optimal-transport

\bibitem{bib45}
P.-E. Sarlin, D. DeTone, T. Malisiewicz, and A. Rabinovich, ``SuperGlue: Learning Feature Matching with Graph Neural Networks'', in \emph{Proc. IEEE/CVF Conf. Comput. Vis. Pattern Recognit.} 4938--4947 (2020). DOI: https://doi.org/10.1109/CVPR42600.2020.00499
\bibitem{bib46}
S. Oron, T. Dekel, T. Xue, W. T. Freeman, and S. Avidan, ``Best-Buddies Similarity---Robust Template Matching Using Mutual Nearest Neighbors'', IEEE Trans. Pattern Anal. Mach. Intell. 40, 1799--1813 (2018). DOI: https://doi.org/10.1109/TPAMI.2017.2737424
\bibitem{bib47}
L. Liu, H. Jiang, P. He, W. Chen, X. Liu, J. Gao, and J. Han, ``On the Variance of the Adaptive Learning Rate and Beyond'', in \emph{Int. Conf. Learn. Represent.} (2020). DOI: https://doi.org/10.48550/arXiv.1908.03265

\bibitem{bib48}
M. A. Fischler and R. C. Bolles, ``Random sample consensus: A paradigm for model fitting with applications to image analysis and automated cartography'', Commun. ACM 24, 381--395 (1981). DOI: https://doi.org/10.1145/358669.358692
\bibitem{bib49}
Y. Han, F. Chen, F. Zeng, C. Lu, Z. Peng, and G. Liu, ``Method for matching online and offline dark-field images of damage sites on large-aperture optical elements'', Chinese patent CN113237888B (June 14, 2022).

\end{thebibliography}
\end{document}